4# Model Uncertainty Quantification for Reliable Deep Vision Structural Health Monitoring

4
Seyed Omid Sajedi & Xiao Liang

*Department of Civil, Structural and Environmental Engineering, University at Buffalo, the State University of New York, Buffalo, NY, United States*

**Abstract:** Computer vision leveraging deep learning has achieved significant success in the last decade. Despite the promising performance of the existing deep models in the recent literature, the extent of models' reliability remains unknown. Structural health monitoring (SHM) is a crucial task for the safety and sustainability of structures, and thus prediction mistakes can have fatal outcomes. This paper proposes Bayesian inference for deep vision SHM models where uncertainty can be quantified using the Monte Carlo dropout sampling. Three independent case studies for cracks, local damage identification, and bridge component detection are investigated using Bayesian inference. Aside from better prediction results, mean class softmax variance and entropy, the two uncertainty metrics, are shown to have good correlations with misclassifications. While the uncertainty metrics can be used to trigger human intervention and potentially improve prediction results, interpretation of uncertainty mask can be challenging. Therefore, surrogate models are introduced to take the uncertainty as input such that the performance can be further boosted. The proposed methodology in this paper can be applied to future deep vision SHM frameworks to incorporate model uncertainty in the inspection processes.

## 1. INTRODUCTION

Visual inspections are an indispensable part of structural health monitoring (SHM). Condition assessments can be carried out periodically as a part of the maintenance program or in near real-time after the extreme events (Sajedi & Liang 2019, 2020). Considering several factors such as time-cost constraints, reliability, and life-safety concerns of human-based inspections, there has been a growing incentive for automation in SHM. Extracting useful information from images is considered as a challenging task, especially in the presence of noisy and complex backgrounds. Deep learning algorithms have shown great promise in dealing with real-world images and more advanced ones are continuously being developed (e.g., Jégou et al. 2017, Chen et al. 2019). This progress has motivated researchers to investigate the potential applications of the machine and deep learning in civil engineering.

There have been efforts to apply these algorithms in predicting the material properties (Rafiei et al. 2017), asphalt surface analysis (Tong et al. 2018), recovering lost sensor data (Oh et al. 2019), etc. More specifically, SHM research has significantly benefited from utilizing deep learning in processing information to assess structural conditions. This input information can be from various sources such as vibration (Abdeljaber et al. 2018, Rafiei & Adeli 2018, Azimi & Pekcan 2019, Khodabandehlou et al. 2019, Sajedi & Liang 2020), acoustic emissions (Ebrahimkhanlou et al. 2019), and strain measurements (Karypidis et al. 2019). While being good indicators of structural damage, acquiring these types of records commonly requires special instrumentations. Hence, their availability is commonly limited to specific case studies, simulations, and lab experiments. This has made the generalization capability of these data-driven models challenging, especially for training deep learning models which requires a substantial number of observations. In contrast, images are relatively more accessible both in terms of quality and quantity. This is partially thanks to camera-equipped drones that can easily capture images that are challenging to obtain by a human inspector (e.g., Liu et al. 2019). Moreover, a significant amount of research is dedicated to convert raw visual records into information that can be utilized for the inspection and monitoring of structures (Spencer et al. 2019).

The vision-based algorithms are generally investigated in two main categories: object detection and damage classification. Identifying the structural components has been done in forms of predicting objects' bounding boxes (Liang 2019) and pixel-wise segmentation of the whole scene (Narazaki et al. 2019). Information obtained from these models can be used for further inspection guidance. The second category (i.e., damage detection) includes identifying various types of damage. Pavement defects and road conditions have been studied using different algorithms including probability generative models and support vector machines (Ai et al. 2018), convolutional neural networks (Maeda et al. 2018, Bang et al. 2019), and recurrent neural networks (Zhang et al. 2017). Identifying cracks has also been the focus of several SHM studies using either bounding boxes (Cha et al. 2017, Xue & Li 2018, Deng et al. 2019) or semantic segmentation (Yang et al. 2018, Sajedi & Liang 2019, Zhang et al. 2019). Other types of structural defects, such as delamination (Cha et al. 2018), cavity (Li et al. 2018, Zhang et al. 2019), fatigue cracks (Hoskere et al. 2018), and efflorescence (Li et al. 2019), or a subset of them are identified using deep learning architectures.

By taking a closer look at this literature review, it is evident that deep learning for visual inspections is becoming more trending among the SHM research community, especially in the past three years. Artificial intelligence may revolutionize SHM research and practice. Nevertheless, certain facts should be kept in mind: proposed models in the literature are data-driven, and thus their performance is highly dependent on the quantity and quality of the training data. While there have been efforts to improve the overall performance, such as using transfer learning and data augmentation (Gao & Mosalam 2018), deep learning algorithms are not *mistake-free*.

Erroneous predictions by vision models have caused fatal accidents in the past where the side of a trailer vehicle on as the sky in a self-driving car (McAllister et al. 2017). We believe that the consequences of misclassifications in structural inspections could be far more severe because SHM investigates structural damage and safety directly. Therefore, it is necessary to have a measure of the model's confidence. This need is further highlighted considering the much smaller size of image datasets for civil engineering applications compared to those of computer science like ImageNet (Deng et al. 2009, Gao & Mosalam 2018).

In this paper, we propose to use the Bayesian inference for SHM vision tasks. This model's uncertainty output, alongside the predictions, can be used to trigger human interventions when the model uncertainty is high (e.g., monitoring damage in a nuclear power plant). The following section provides a brief underlying theory for Bayesian segmentation networks. Later in the paper, examples from three independent case-studies are presented to show that the uncertainty metrics can correlate well when there are mistakes in the models (sections 3-5). In each section, the Bayesian inference is also compared with the benchmark models in terms of several performance metrics. Despite the superior robustness of Bayesian models, we propose a novel surrogate approach. This extension of the framework can automatically incorporate the uncertainty output and further enhance the performance of our models.

## 2. DEEP LEARNING FRAMEWORK WITH BAYESIAN INFERENCE

This section gives a brief theoretical background on Bayesian inference and how model uncertainty can be quantified in deep segmentation models. Later, the deep learning architecture used in this paper is explained.

### 2.1. Inference and Uncertainty Quantification

Uncertainty should be a natural part of any predictive system's output. Knowing the confidence with which we can trust the damage diagnosis output is essential for decision making. Bayesian probability theory (Koller & Friedman 2009) offers a mathematically grounded framework to reason about model uncertainty but usually comes with a prohibitive computation cost, especially for real-time implementation in computer vision (Gal & Ghahramani 2016). The use of dropout (and its variants) in neural networks can be interpreted as a Bayesian approximation of a well-known probabilistic model: The Gaussian process (Gal & Ghahramani 2016). With a prior probability, this operator randomly sets a fraction of input elements into zero as a way to reduce overfitting while training a neural network. The standard dropout uses the weight averaging technique (Srivastava et al. 2014) at test time such that the deep learning models produce deterministic output. Leveraging Bayesian inference, a probabilistic interpretation of a deep convolutional encoder-decoder network will be developed by inferring distribution over the networks' weights. Variational inference (Graves 2011) will be used to approximate the model's intractable posterior distribution.

Training the network, i.e., minimizing the cross-entropy loss objective function, has the effect of minimizing the Kullback-Leibler divergence between this approximating distribution and the full posterior (Gal & Ghahramani 2015) such that the approximating distribution will be learned. Specifically, the dropout will be used at test time to approximate the model's intractable posterior distribution by imposing a Bernoulli distribution (Gal & Ghahramani 2015) across the network's weights. This can be achieved by sampling the network with randomly dropped out units, which can be considered as Monte Carlo sampling from the posterior distribution over the network.

Based on the brief theoretical overview, Bayesian inference can be conveniently implemented for deep vision structural inspections. In pixel-wise image segmentation, a regular prediction output ($y$) is a tensor of shape (*height*, *width*, $N_b$) where the last channel refers to the softmax probabilities of class $i$ ($S_i$, $i \in \{1, 2, ..., N_b\}$). Feeding a specific input image to a deep learning model with standard inference will result in deterministic $S_i$ values, the maximum of which is considered as the decision probability. In contrast, activating dropout layers at the inference time yields a prediction output where class probabilities can be deemed to be random variables. For a single observation, $N_s$ Monte Carlo samples can be drawn to form a stacked output tensor with the shape of (*height*, *width*, $N_b$, $N_s$). The expected value (mean) of $S_i$ samples are then used for the final inference. Therefore, the probability of an observation (*y*) belonging to the class *i* can be expressed using Equation (1) (Gal 2016). This equation highlights the key difference between the Bayesian and standard inferences. The later directly uses constant $S_i$ values that are obtained from a neural network without dropout while the Bayesian inference utilizes the mean of the $N_s$ samples.

$$p(y = i \mid \mathbf{X}, \mathbf{Y}) \approx \mathrm{E}(S_i) = \frac{1}{N_s} \sum_{n=1}^{N_s} S_i^n \qquad (1)$$

where $\mathbf{X}$ and $\mathbf{Y}$ are, respectively, the input (image) and segmentation masks.

Two metrics are introduced to quantify model uncertainty. Kendall & Gal (2017) propose Entropy (*H*) as a measure of model epistemic uncertainty:

$$H(p) = \sum_{i=1}^{N_b} -p_i \log(p_i) \qquad (2)$$

Moreover, class softmax variance (*CSV*) can also be used as an indicator of the model uncertainty for each class (Kendall et al. 2015). This metric is obtained by taking the sample variance of Monte Carlo samples ($S_i$'s) with respect to each class:

$$CSV_i = \frac{1}{N_s - 1} \sum_{n=1}^{N_s} [S_i^n - \mathrm{E}(S_i)]^2 \qquad (3)$$

As the second overall measure of model uncertainty, the mean of the per-class softmax variances (*MCSV*) is used. In segmentation, Entropy and *MCSV* can be obtained for all pixels. Therefore, the two measures of uncertainty are presented for all pixels in the form of a 2D tensor with the same dimension as that of the final segmentation mask.

It should be emphasized that there is a fundamental difference between softmax probabilities and model confidence (Gal & Ghahramani 2016). In other words, they are not necessarily correlated, e.g., a predicted label with a high softmax probability may have high model uncertainty.

**2.2. Deep Learning Architecture**

In this paper, Fully Convolutional (FC) DenseNet is leveraged to perform Bayesian inference. This neural network combines the underlying idea of U-Net (Ronneberger et al. 2015) and DenseNets (Huang et al. 2017). The architecture has achieved state-of-the-art performance on benchmark datasets for urban scene segmentation (Jégou et al. 2017) and is considered among the most successful ones for the given task. Similar to most image segmentation algorithms, FC-DenseNet automatically extracts features while reducing the spatial resolution of feature maps by performing multiple pooling operations in the downsampling path. The spatial resolution of the input is later recovered in the upsampling path. The two paths are connected with a bottleneck in between. What makes this architecture unique is the presence of sophisticated connectivity patterns where input from the previous layers is concatenated with the extracted features maps. Moreover, several skip connections help to recover fine-grained information in the upsampling process.

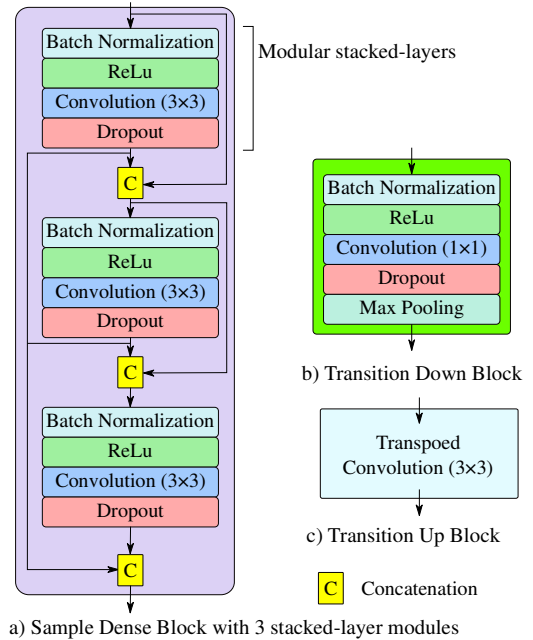

a) Sample Dense Block with 3 stacked-layer modules
Figure 1. FC-DenseNet building blocks

The building blocks of FC-DenseNet are Dense Blocks (DBs), Transition up (TU), and Transition down (TD) units. The details of each unit are shown in Figure 1. Dense blocks are comprised of modular stacked-layers. Each module contains a sequence of batch normalization, ReLu activation, convolution, and dropout layers (Figure 1.a). The output of each module is then concatenated with its input. The first convolution will extract 48 filters while the number of filters inside DB convolutions (growth rate) is 16. The TD units are similar to a modular stacked-layer while it has an additional max-pooling operation to reduce the spatial resolution (Figure 1.b). The number of filters in their corresponding 1×1 convolution is equal to the final output of the previous dense block. Finally, TU units are simply transposed 3×3 convolution (Long et al. 2015).

In this paper, we investigate three different case studies. The first dataset deals with binary crack detection. The second one is dedicated to localizing damage. Finally, a multi-class task of bridge component recognition is investigated as the third case-study. Given the different complexity, the original architecture is tailored for computational efficiency. Therefore, the number of modular stacked units inside the dense blocks are modified for the binary problems. The details are given in Table 1.

The general structure of the adopted deep learning architecture for Bayesian inference is illustrated in Figure 2. Recall that the dropout layers are active in both training and inference phases. For a single input image, a number of Monte Carlo dropout samples are drawn to generate a bin of softmax probabilities, form which the prediction and uncertainty metrics can be obtained. This bin is generated assuming a total of 50 Monte Carlo samples with a 50% dropout probability.

Table 1. Number of modular stacked layers in dense blocks

| Block ID | Models 1&2 | Model 3 |
|---|---|---|
| DB-1, DB-11 | 2 | 4 |
| DB-2, DB-10 | 3 | 5 |
| DB-3, DB-9 | 4 | 7 |
| DB-4, DB-8 | 5 | 10 |
| DB-5, DB-7 | 6 | 12 |
| DB-6 (Bottle neck) | 8 | 15 |

All three models are trained using Keras API (Chollet 2015) with a batch size of 2 on computers equipped with NVIDIA GTX 1070 and GTX 1080, each with 8 GB of memory. Observations are randomly shuffled and 80% are considered for training and the remaining are used as the test set. Moreover, 20% of training observations are held out as a validation set. The model with the lowest validation loss is selected for testing. It should be noted that Bayesian inference is used for both validation and test sets.

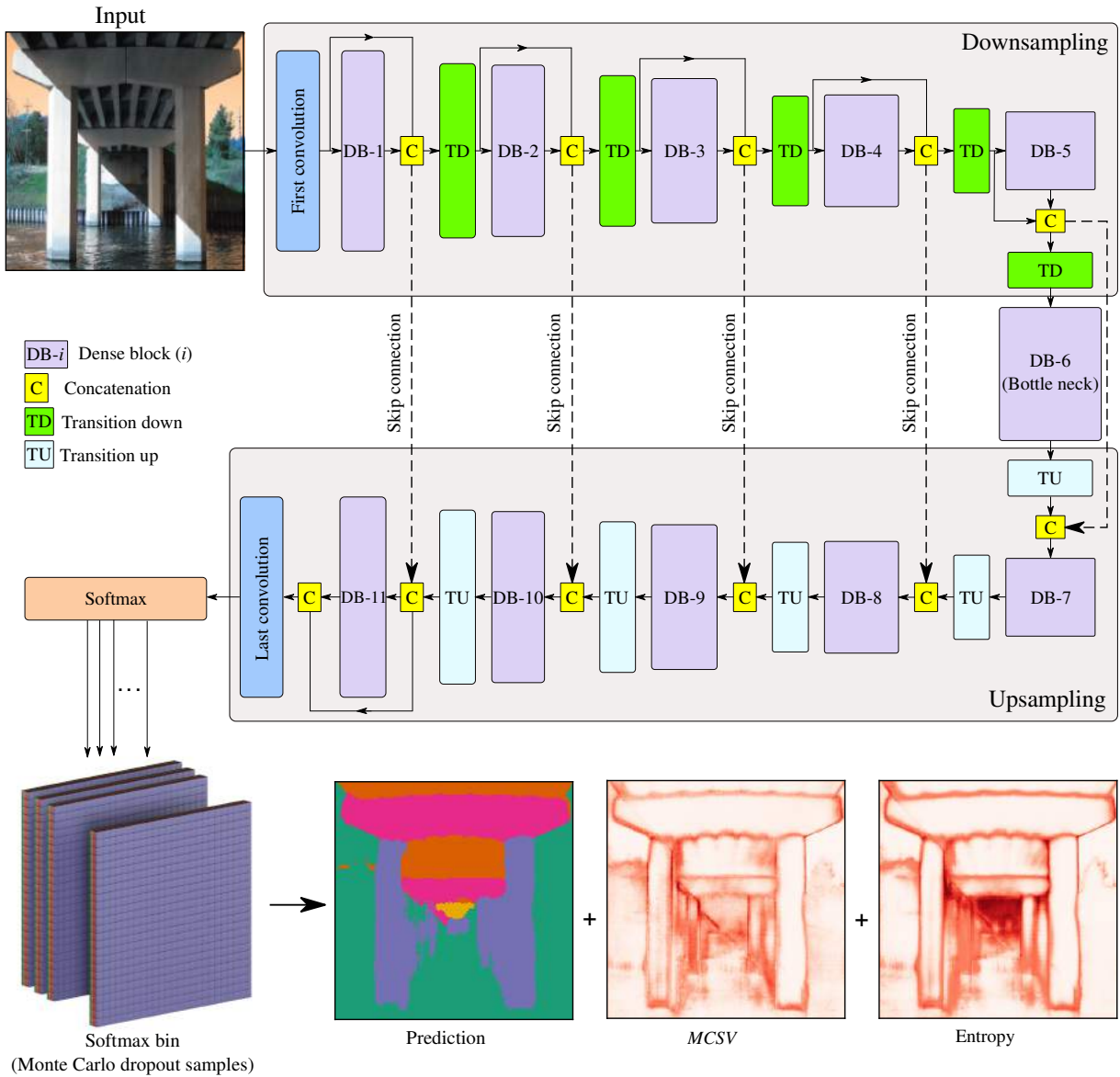

Figure 2. Bayesian Inference with FC-DenseNet architecture (parts of figure are inspired by Jégou et al. 2017)

Based on our experiments, Nadam optimizer with exponential weight decay (per epochs) yields a faster convergence with a fewer number of epochs. L2 regularization (Goodfellow et al. 2016) is also used for better numerical stability and less overfitting. All models are trained with a maximum of 200 epochs and an early stopping of 15 epochs based on the validation loss. A learning rate of 1.0e-4 and an exponential decay rate of 0.9996 are also used in training.

## 3. BINARY CRACK DETECTION (I)

Binary segmentation of cracks is one of the well-studied areas for automated SHM. In this task, the crack forest dataset (Shi et al. 2016) is utilized. The dataset includes 118 color images with a resolution of 320×480 pixels. The images are taken by a camera-equipped smartphone with nearly constant photography setup (e.g., distance, exposure, aperture). The dataset reflects information on surface road conditions. The ground truth is processed to include a binary mask with two classes: crack and background.

Crack segmentation is a challenging vision problem because of the significant class imbalance. Images that reflect a cracked surface are commonly dominated by background pixels. In the current dataset, less than 2% of the ground truth pixels are labeled as crack. As a result, global accuracy (*GA*) is not a proper metric for this task. For example, if one labels all pixels as background, *GA* will be approximately 98% while no cracks are successfully detected.

The original implementation of FC-DenseNet utilizes uniform weights (UW) in the loss function and maximum a-posteriori (MAP) decision rule, which takes the maximum of softmax probabilities. Eigen & Fergus (2015) propose a median frequency weight (MFW) assignment to help increase the mean class accuracy (*MCA*). In more recent work, Chan et al. (2019) propose modifying the decision rule instead of adjusting the observation weights in the loss function. This goal is achieved by weighing the posterior class probabilities of crack and background classes with their inverse frequency in the training data. The decision rule for this type of inference is called the maximum likelihood (ML) as an alternative to MAP.

Considering the two previous methods and the original setup in the FC-DenseNet, three different strategies of UW-MAP, UW-ML, and MFW-MAP are investigated. The first term relates to the way that weights are balanced and the second indicates the decision rule to assign a label to each pixel. For example, UW-ML means that the loss function is minimized assuming uniform weights for both crack and background pixels. Moreover, the ML decision rule is used to assign a label to each pixel.

In all case studies, two sets of models are trained. A benchmark model without dropout is used to compare the results with the Bayesian model. For each of the three strategies mentioned earlier, the two models are compared with each other. A summary of the results for a total of 6 combinations is given in Table 2. *GA* and *MCA* are obtained for the test dataset. In addition, F1-score, precision, recall, and intersection over the union (IoU) are also measured for the cracks. Note that these metrics are almost perfect for the background class and of less interest for this dataset.

The information in Table 2 can be analyzed from different perspectives. Compared with the benchmark, it is evident that the Bayesian models achieve superior performance for all criteria. As shown in Figure 3, some crack patterns, that are missed by the benchmark models, are properly identified using Bayesian inference. Beyond the numerical metrics, this observation is also visually evident by comparing each pair of images given the same strategy. This improvement can also be highlighted under the MFW-MAP column as the noisy misclassifications are considerably reduced after Bayesian inference (e.g., Figure 3.3).

It is also of interest to select one of the models with respect to the three strategies that are used. For example, the UW-MAP model has the highest GA. Nevertheless, by looking at the examples provided in Figure 3, it is clear that this approach (as in the original FC-DenseNet) does not perform well in this highly imbalanced dataset. Several crack patterns are missed despite having the highest global accuracy among the three. In contrast, the MFW-MAP captures most cracks (highest recall among the three strategies) while being the most conservative approach (with the lowest precision).

Table 2. Testing performance of 6 different combinations for the Crack Forest dataset

| Model ID | GA | | | MCA | | | Crack F1 score | | |
|---|---|---|---|---|---|---|---|---|---|
| | UW-MAP | UW-ML | MFW-MAP | UW-MAP | UW_ML | MFW-MAP | UW-MAP | UW-ML | MFW-MAP |
| Benchmark | 98.57 | 98.35 | 95.87 | 77.05 | 89.76 | 89.88 | 60.02 | 65.83 | 44.40 |
| Bayesian models | **98.76** | **98.60** | **96.83** | **81.48** | **93.37** | **93.39** | **66.93** | **71.18** | **52.74** |
| | Crack IoU | | | Crack precision | | | Crack recall | | |
| | UW-MAP | UW-ML | MFW-MAP | UW-MAP | UW-ML | MFW-MAP | UW-MAP | UW-ML | MFW-MAP |
| Benchmark | 42.80 | 49.06 | 28.54 | 66.57 | 55.53 | 30.23 | 54.65 | 80.82 | 83.63 |
| Bayesian models | **50.30** | **55.26** | **35.82** | **70.77** | **59.79** | **37.34** | **63.49** | **87.94** | **89.81** |

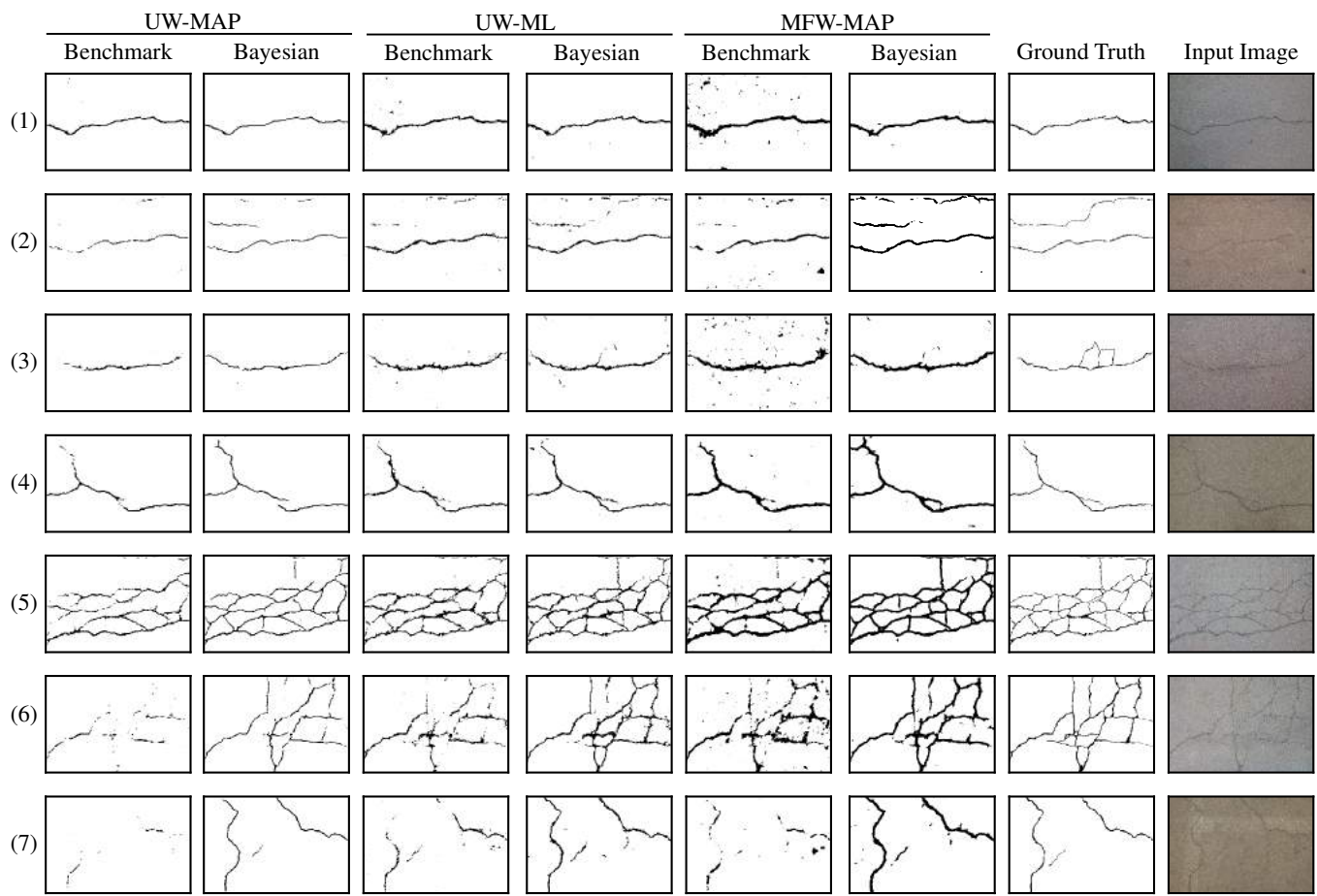

Figure 3. Sample testing observations for binary crack segmentation: comparisons between the benchmark and Bayesian inference models

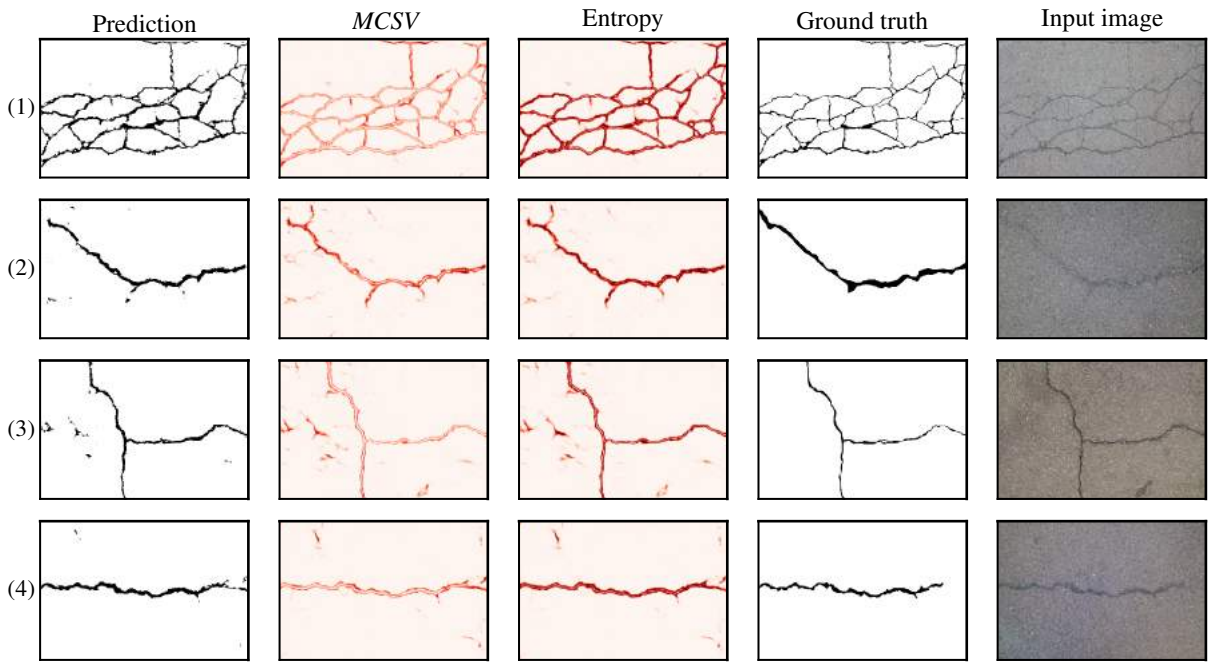

Figure 4. Model uncertainty measures using Bayesian inference

Table 3. Testing performance metrics on the second dataset

|  | Background | | Damage | | Mean value | |
|---|---|---|---|---|---|---|
|  | Benchmark | Bayesian models | Benchmark | Bayesian models | Benchmark | Bayesian models |
| F1-score (%) | 92.54 | **93.99** | 64.93 | **70.06** | 78.74 | **82.03** |
| Precision (%) | 92.94 | **93.46** | 63.64 | **72.08** | 78.29 | **82.77** |
| Class Accuracy* (%) | 92.14 | **94.52** | 66.26 | **68.14** | 79.20 | **81.33** |
| IoU (%) | 86.12 | **88.67** | 48.07 | **53.92** | 67.10 | **71.30** |

* Recall is another term for the class accuracy

Considering all metrics shown in Table 2, the UW-MAP and MFW-MAP are the two extreme cases where the model's performance are unsatisfied in terms of either precision or recall. Since both metrics represent qualities that matter to a decision-maker, the Bayesian UW-ML is deemed as the best among the others. It shows a reasonable balance between precision and recall, which is also evident by its highest F1 score.

Aside from the improvements in the prediction results, the main advantage of Bayesian inference for vision SHM is the uncertainty output. The two model's uncertainty measures (*MCSV* and entropy) are illustrated in Figure 4. Sample testing observations in this figure are from the UW-ML strategy described earlier. By taking a closer look at the distribution of uncertainties, it can be observed that model uncertainty is relatively high at the crack boundaries. Furthermore, an important take from this figure is that the model uncertainty correlates well with misclassifications. For example, the model shows high uncertainty in classifying stains (as in Figure 4.3) or identifying the background noises as crack. Moreover, both metrics are good indicators of model uncertainty while being different in values and distribution. Considering the correlation between mistakes and model uncertainty, entropy is more conservative.

## 4. DAMAGE LOCALIZATION (II)

The second case study is dedicated to localizing the presence of damage. The dataset is obtained from Liang (2019) which includes 436 images with a resolution of 320×480 pixels. The observations are obtained from two main sources: laboratory experiments on reinforced concrete columns and post-earthquake damage records of bridges. Unlike crack forest with close-up figures and minimum background noise, the images in this dataset commonly have complex backgrounds. This complexity is even amplified for lab experiments where the structural elements are commonly covered with testing instruments. This dataset is also labeled with a binary mask of two classes: damage and background. The performance metrics for this task are presented in Table 3. It should be noted that the UW-MAP strategy is used for this case study.

Similar to the crack segmentation, the Bayesian model is compared with a corresponding benchmark. The two models are the same in terms of architecture and training hyperparameters except for the existence of dropout layers in training and inference. For all the considered metrics, significant improvements are observed.

Sample testing observations of this dataset are given in Figure 5. The segmentation results are in good agreement with the ground truths for the majority of observations (including images taken from the laboratory experiments). However, there exist some examples that show poor predictions. For example, in Figure 5.19, the majority of the damaged area is missed in the prediction mask. By comparing the ground truth with the uncertainty metrics, one can see that the misclassified damage area is associated with high model uncertainty. This is an example illustrating the importance of model uncertainty output, which can be used to trigger human intervention. In this case, an inspector can be warned to re-evaluate the condition of the structure.

## 5. BRIDGE COMPONENT RECOGNITION (III)

For the third case-study, we propose an example that deals with object recognition rather than damage diagnosis. As mentioned in the literature review, automatic identification of different objects (e.g., structural components) is crucial, especially in guiding unmanned aerial vehicles (UAVs) for inspection (e.g., Liang et al. 2018, Zheng et al. 2019). Bridges are important parts of the civil infrastructure that will be benefited from such camera-equipped UAV inspections. To this end, the third dataset is comprised of 236 images of highway bridges. While originally used to identify bridge columns with bounding boxes (Liang 2019), bridge components are pixel-wise labeled by the authors using MATLAB's image labeler toolbox (Mathworks 2019) for semantic segmentation.

Table 4. Testing performance in bridge component detection

|  | F1-score (%) | | Precision (%) | | Class Accuracy (%) | | IoU (%) | |
|---|---|---|---|---|---|---|---|---|
|  | Benchmark | Bayesian | Benchmark | Bayesian | Benchmark | Bayesian | Benchmark | Bayesian |
| Background | 89.83 | **92.59** | 84.84 | **89.74** | 95.44 | **95.62** | 81.54 | **86.17** |
| Superstructure | 82.85 | **86.75** | 78.37 | **82.10** | 87.88 | **91.96** | 70.73 | **76.60** |
| Column | 82.21 | **87.23** | 77.99 | **89.19** | **86.90** | 85.36 | 69.80 | **77.35** |
| Cap beam | 31.89 | **71.70** | **88.05** | 81.73 | 19.47 | **63.87** | 18.98 | **56.09** |
| Foundation | 3.51 | **28.99** | **100.00** | 74.64 | 1.79 | **17.99** | 1.79 | **16.98** |
| Abutment | 35.27 | **62.19** | 92.57 | **93.04** | 22.13 | **46.70** | 21.75 | **44.81** |
| Mean value | 54.26 | **71.58** | **86.97** | 85.07 | 52.27 | **66.92** | 44.10 | **59.67** |

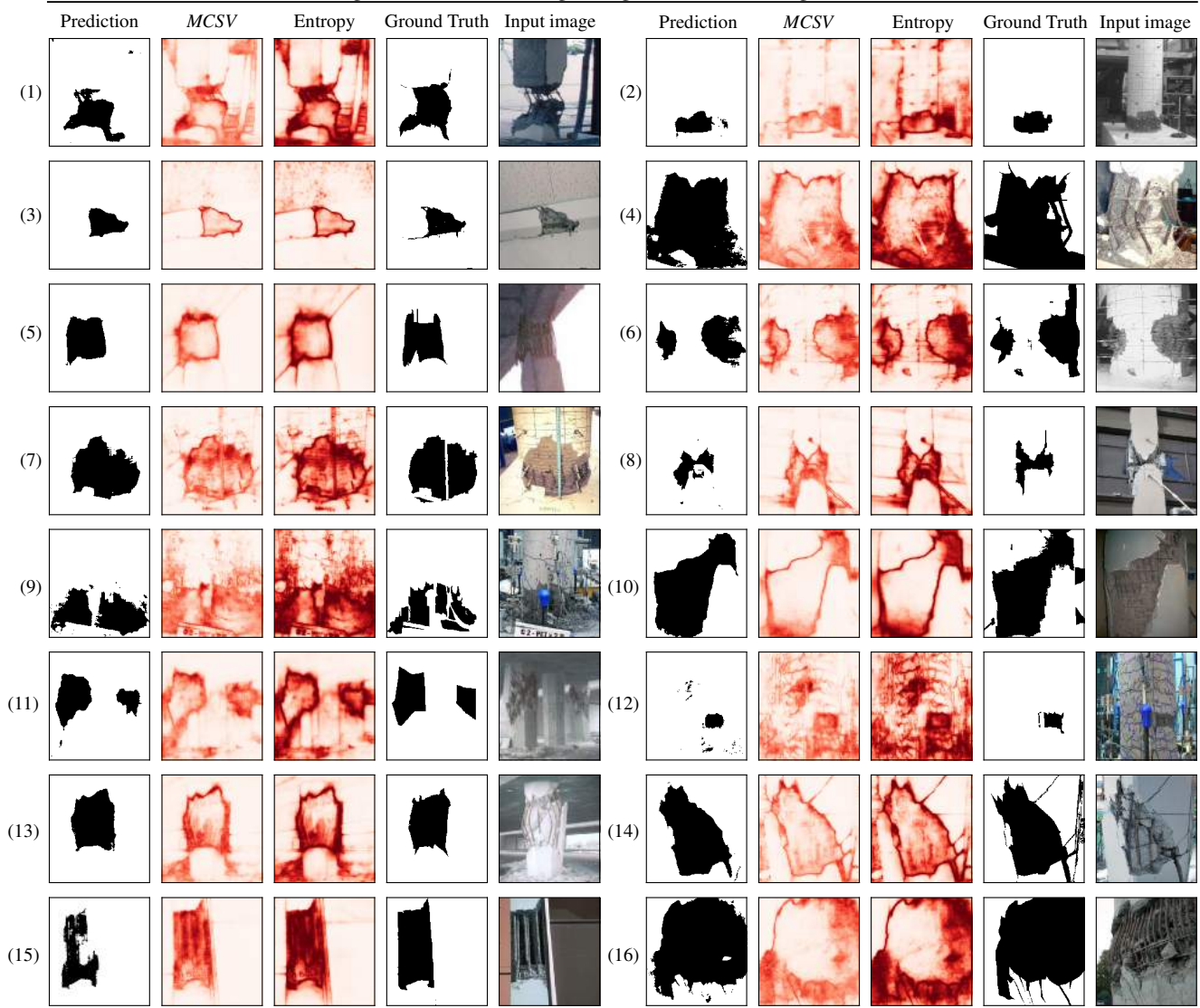

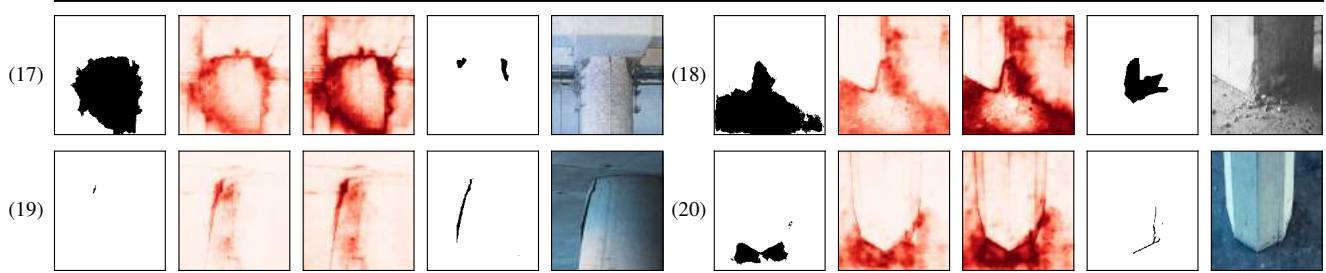

Figure 5. Sample testing observations for damage localization using Bayesian inference

Different from the first two case studies, this one deals with a multi-class segmentation problem. The assumed structural components of interest for inspection are the bridge column (pier), cap beam, abutment, foundation, and superstructure. All the other miscellaneous objects in the scene are labeled as background. An important point about this dataset is that the abutment and foundation are usually buried inside the ground or soil for stabilization and are commonly less exposed. As a result, the frequency of pixel observations for the foundation and abutment is considerably less than that of the other classes.

The UW-MAP strategy is used for training and inference and a summary of performance metrics is presented in Table 4. It is evident that, compared with other components, the less frequent classes of abutment and foundation are detected with less accuracy. Similar to the previous two case studies, the superior performance of the Bayesian inference over the benchmark model is evident. Another observation from Table 4 is that the precision for the classes of cap beam and foundation is higher in the benchmark. It should be emphasized that this high precision is accompanied by a significantly smaller recall compared with the Bayesian model. In this case, high precision does not imply better performance. For example, only 1.79% of the pixels are correctly captured as a foundation by the benchmark model with a 100% precision.

Sample testing observations for bridge component segmentation are presented in Figure 6. Similar to the other datasets, the presence of misclassification has a good correlation with the model uncertainty. Yet, there exist cases where the uncertainty is low while the segmentations are not accurate. For example, Figure 6.12 illustrates a column in a laboratory setup. The column is properly identified but the upper region of the image is misclassified as the superstructure. While this is an unlikely scene to happen for an outdoor inspection, it is important to be aware of the cases where the model uncertainty does not reflect mistakes. In this example, the model shows confidence that the superstructure is the closest prediction for the roof of a laboratory compared with other possible classes in the dataset. One potential reason for this type of inconsistency is the fact that no knowledge about the existence of the laboratory roof is introduced in data labeling. It is important to consider the limitations of supervised learning caused by the lack of sufficient information in the dataset. Such inconsistencies can be alleviated by enlarging the training set for such observations and potentially increasing the number of possible classes.

In the two previous binary segmentations, despite the difference in the relative values, the two uncertainty metrics follow a similar distribution. In this multi-class dataset, noticeable differences between the distribution patterns of entropy and *MCSV* are observed. Figure 6.1 is an example that highlights this difference where some pixels have relatively high entropy but low *MCSV* and vice-versa. Observations like this will make the incorporation of uncertainty for the decision-making challenging. It is difficult or impossible for a human to simultaneously interpret the entropy and *MCSV* for all pixels and modify the decision accordingly. The remainder of the paper elaborates on such complications and proposes a potential solution to benefit from the uncertainty output.

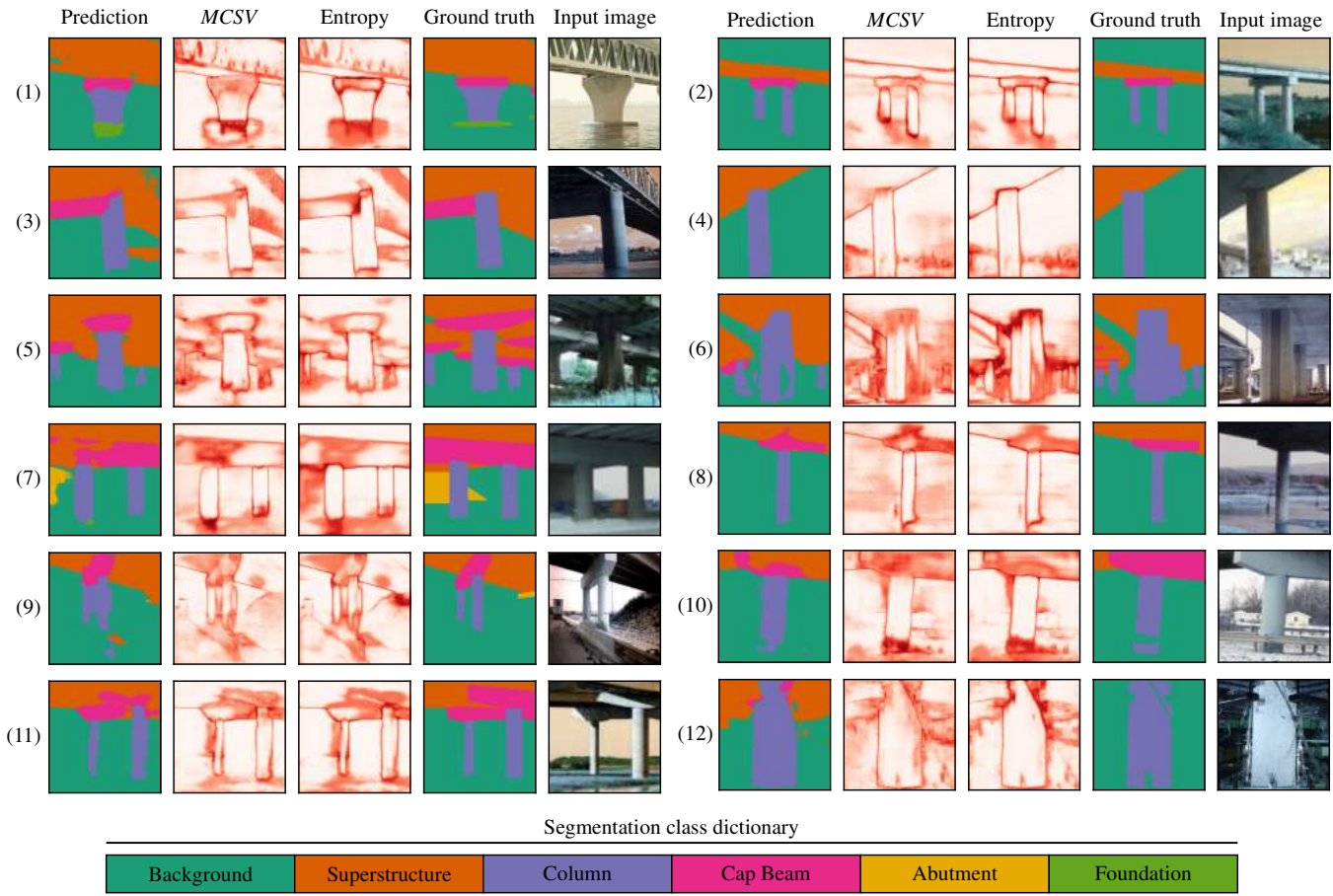

Figure 6. Sample testing observations for bridge component recognition using Bayesian inference

## 6. SURROGATE UNCERTAINTY-ASSISTED SEGMENTATION

In sections 3-5, we have showed that Bayesian inference has two potential advantages for deep vision SHM. First, it improves the overall performance of a model. Second, Bayesian deep neural networks provides a model uncertainty output. The importance of this output is crucial for the condition assessment of critical structures such as nuclear facilities where missing a simple defect may lead to catastrophic outcomes. Quantification of uncertainty helps further minimize the risk of using data-driven models for safety-critical inspections. Being informed of the lack of confidence in a model's predictions, a decision-maker can further explore the scene.

Earlier in the paper, several examples were provided where the uncertainty mask correlates well with the boundaries in the regions of interest (i.e., cracks, local damage, and bridge components). Moreover, compared with correct predictions, a considerable portion of misclassified pixels is shown to have a relatively high value of model uncertainty.

These correlations visually make sense because one can perform a side-by-side comparison between the uncertainty mask and the ground truth. High uncertainty in the output may be sufficient to trigger an alarm for further inspections. The next and more important step is to exploit the existing uncertainty masks and correct the prediction results accordingly.

When exploring the field or performing an additional expert analysis is not possible, model uncertainty is the only source of information to potentially enhance the predictions. In such circumstances, a human operator may also face challenges in how to utilize the *MCSV* or entropy for a better estimate of real conditions. For example, this utilization is very complicated (or even impossible) for a human if the convolutional neural networks are used for regression tasks such as depth estimation of pixels.

Another challenge in this regard is the real-time availability of predictions which will be substantially limited when human intervention is required. The majority of the proposed frameworks are designed/optimized to partially automate the SHM process. Assigning a human operator to monitor the uncertainty output and to modify the predictions is not feasible for certain SHM tasks that require processed information in real-time. For example, automatic guidance of a UAVs for inspections will not be automatic anymore if it is constantly interrupted by manual instructions.

In this section, we propose a way regarding *how* to benefit from the uncertainty mask without human intervention. This goal is achieved by introducing a second data-driven model to automatically interpret the *MCSV* and entropy masks. We call it a *surrogate* model because identifying uncertainty rules for thousands of pixels, that may or may not be correlated, is not directly possible in an image. After training the original model and calibrating the learnable parameters, the uncertainty output is available. Therefore, it is desired to construct a second input by concatenating the original image with the mean softmax probabilities (E($S_i$)), class softmax variances ($CSV_i$), and entropy in stacked channels (see Figure 7). The surrogate model is trained in a supervised manner with the second input and the same labeled pixels as the output. Different from the initial Bayesian segmentation, the surrogate model has access to uncertainty output. It is trained to learn the underlying mapping between the second input (image data combined with early prediction uncertainties) and the ground truth.

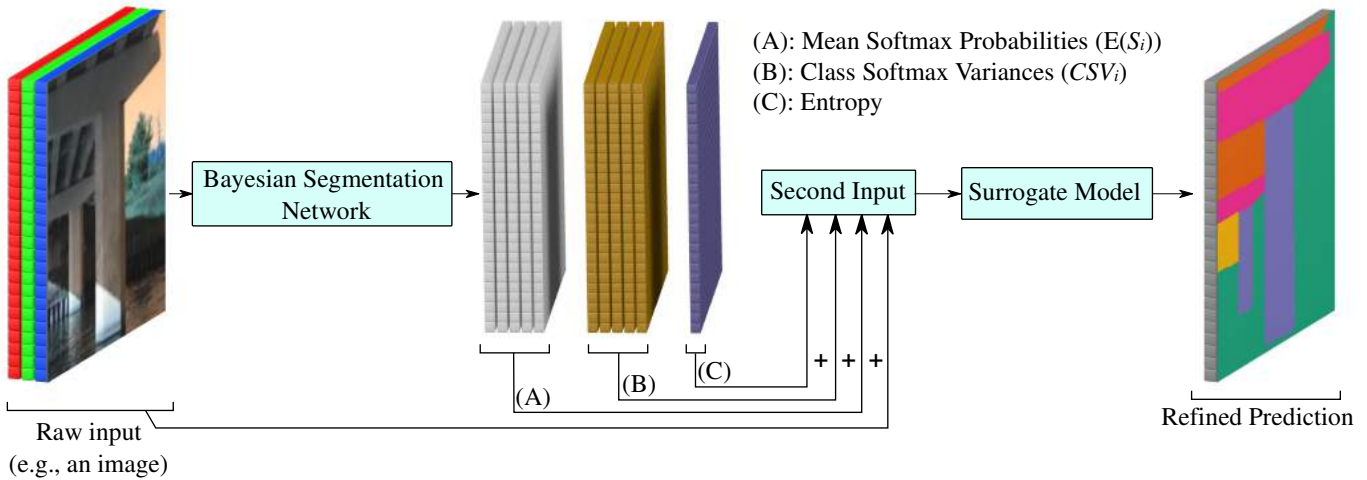

Figure 7. Surrogate uncertainty-assisted model with the second input

Table 5. Comparison of test set performance between the surrogate models and initial segmentation (all values are in percentage)

|  | Crack segmentation (I) | | | Damage segmentation (II) | | |
|---|---|---|---|---|---|---|
|  | Background | damage | Average | Background | damage | Average |
|  | F1 score | | | F1 score | | |
| Bayesian Inference | 99.28 | 71.18 | 85.23 | 93.99 | 70.06 | 82.025 |
| Surrogate model | 99.36 | 72.62 | 85.99 | 94.19 | 71.62 | 82.905 |
| Difference | **+0.08** | **+1.44** | **+0.76** | **+0.20** | **+1.56** | **+0.88** |
|  | Class accuracy | | | Class accuracy | | |
| Bayesian Inference | 98.81 | 87.93 | 93.37 | 94.52 | 68.15 | 81.34 |
| Surrogate model | 99.06 | 83.49 | 91.28 | 94.40 | 70.83 | 82.62 |
| Difference | **+0.25** | **-4.44** | **-2.09** | **-0.12** | **+2.68** | **+1.28** |
|  | Precision | | | Precision | | |
| Bayesian Inference | 99.76 | 59.79 | 79.78 | 93.47 | 72.08 | 82.78 |
| Surrogate model | 99.66 | 64.25 | 81.96 | 93.98 | 72.41 | 83.20 |
| Difference | **-0.10** | **+4.46** | **+2.18** | **+0.51** | **+0.33** | **+0.42** |
|  | IoU | | | IoU | | |
| Bayesian Inference | 98.57 | 55.26 | 76.915 | 88.67 | 53.92 | 71.30 |
| Surrogate model | 98.74 | 57.02 | 77.88 | 89.01 | 55.78 | 72.40 |
| Difference | **+0.17** | **+1.76** | **+0.97** | **+0.34** | **+1.86** | **+1.10** |

|  | Bridge component recognition (III) | | | | | | |
|---|---|---|---|---|---|---|---|
|  | Background | Superstructure | Column | Cap beam | Foundation | Abutment | Average |
|  | F1 score | | | | | | |
| Bayesian Inference | 92.59 | 86.75 | 87.23 | 71.7 | 28.99 | 62.19 | 71.58 |
| Surrogate model | 92.27 | 87.89 | 87.58 | 75.86 | 32.81 | 66.68 | 73.85 |
| Difference | **-0.32** | **+1.14** | **+0.35** | **+4.16** | **+3.82** | **+4.49** | **+2.27** |
|  | Class accuracy | | | | | | |
| Bayesian Inference | 95.62 | 91.96 | 85.36 | 63.87 | 17.99 | 46.7 | 66.92 |
| Surrogate model | 93.98 | 90.69 | 85.18 | 71.68 | 24.62 | 64.67 | 71.80 |
| Difference | **-1.64** | **-1.27** | **-0.18** | **+7.81** | **+6.63** | **+17.97** | **+4.89** |
|  | Class precision | | | | | | |
| Bayesian Inference | 89.74 | 82.1 | 89.19 | 81.73 | 74.64 | 93.04 | 85.07 |
| Surrogate model | 90.63 | 85.25 | 90.12 | 80.54 | 49.11 | 68.81 | 77.41 |
| Difference | **+0.89** | **+3.15** | **+0.93** | **-1.19** | **-25.53** | **-24.23** | **-7.66** |
|  | IoU | | | | | | |
| Initial segmentation | 86.17 | 76.6 | 77.35 | 56.09 | 16.98 | 44.81 | 59.67 |
| Surrogate model | 85.64 | 78.38 | 77.87 | 61.08 | 19.61 | 50.07 | 62.11 |
| Difference | **-0.53** | **+1.78** | **+0.52** | **+4.99** | **+2.63** | **+5.26** | **+2.44** |

In what follows, it is shown that the proposed approach yields an improved segmentation for the previous case-studies. For the sake of comparison, we use the same architecture of the segmentation network (except the input layer) to train the surrogate model. The only difference is the construction of uncertainty input data. This helps to ensure that the improvement in the results is solely due to the proposed novel input. Therefore, factors such as a deeper architecture or change in the selection of hyperparameters are not a potential reason for the refined predictions. Given this similarity, the second model is also Bayesian.

The comparisons of testing performance between the initial Bayesian inference and the corresponding surrogate model are presented in Table 5 for the three case studies. It is observed that the surrogate models result in overall better performance. For all three datasets, IoU, F1 score, and class accuracy are improved. The difference between the two models is better highlighted in the third case study which has more classes and thus is more complex. This improvement is more significant for the bridge components that are less frequent and thus more difficult to recognize (i.e., abutment and foundation classes). A slight decay is observed in the prediction accuracy of more frequent classes, such as background and deck, while on average, the surrogate models improve the results. It should be also noted that the third model correctly captures a relatively larger number of pixels for the foundation and abutment. Therefore, while the precision is reduced, the class accuracies have significantly increased. For the crack segmentation algorithm, the overall metrics IoU, F1 score, and precision are improved. Although, there is a decay in class accuracy for the cracks. An illustrated comparison between the two models is presented for the second and third case study in Figure 8.

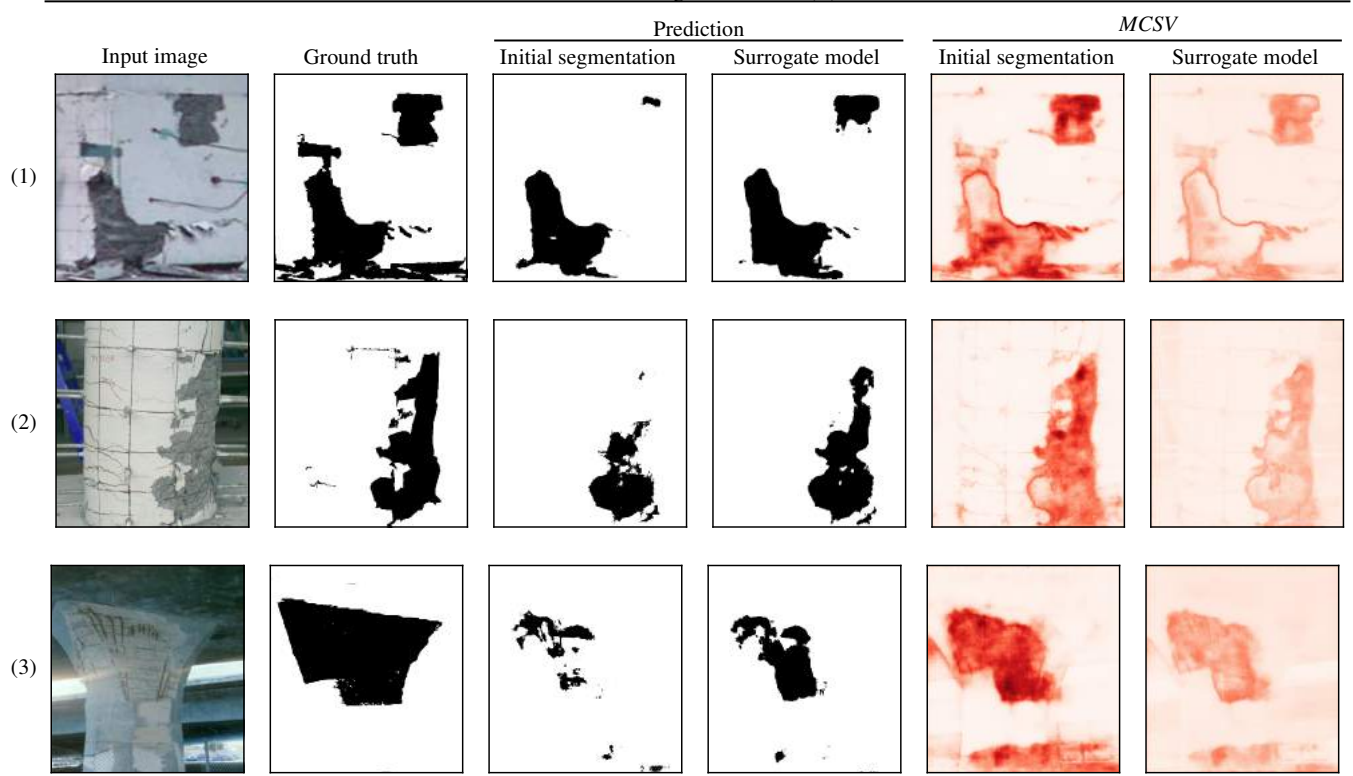

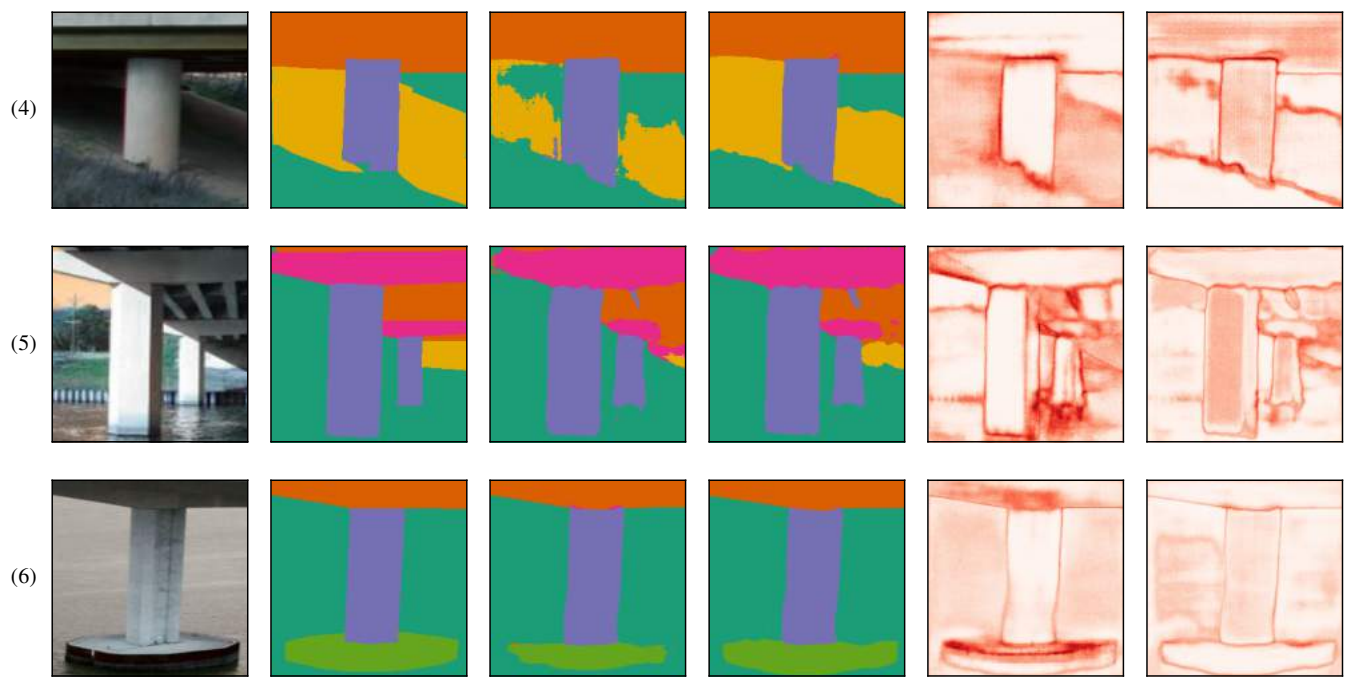

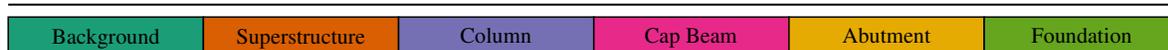

Figure 8. Visual comparison between the initial and surrogate segmentation results

As mentioned earlier, the second model is Bayesian and hence, also outputs uncertainty. Hence, a side by side comparison between the two *MCSV* masks is also presented in Figure 8. The color scale is normalized based on the maximum value of both initial and surrogate masks such that the intensities are comparable. It is clear that the surrogate model has much less uncertainty.

## 7. CONCLUSION

Recent advances in computer vision and deep learning have had a major impact on vision-based structural inspections. Considering an ever-growing interest in the application of deep learning models for structural health monitoring, it is necessary to develop a framework that quantifies the model's confidence. Model uncertainty can be used to alert the decision-maker when the reliability of predictions is questionable. In this paper, we propose a Bayesian deep learning framework for vision-based structural inspections, and its superiority is demonstrated by three different case studies.

In the first case study, 6 different strategies are investigated where Bayesian UW-ML is selected as the best approach for crack detection. A binary damage localization and a multi-class bridge component recognition problem are further investigated. All three case studies show that Bayesian inference provides more robust predictions leveraging Monte Carlo dropout sampling.

Along with the greater robustness, the Bayesian vision models provide the decision-maker with uncertainty measures. It is shown that *MCSV* and entropy are good estimates of model uncertainty. These metrics show good correlations with misclassifications.

The quantified measures of uncertainty can be useful to trigger human interventions when the model confidence is low. This intervention could be challenging or even impossible. To incorporate the uncertainty output as a part of automation, the surrogate uncertainty-assisted model is proposed to further improve the predictions and boost the model's performance.

The methodology proposed in this paper can be utilized to equip the existing vision models for inspection and monitoring with tools that can quantify and output model uncertainty. In the absence of huge datasets for training, Bayesian inference can be an effective tool to make visual inspections using deep vision models more reliable. While we utilize the surrogate models to improve the original image segmentation results, they can be tailored to different needs and enhance the prediction results of other vision-based SHM applications.

## 8. REFERENCES

Abdeljaber, O., O. Avci, M. S. Kiranyaz, B. Boashash, H. Sodano and D. Inman (2018). "1-D CNNs for structural damage detection: verification on a structural health monitoring benchmark data." Neurocomputing **275**: 1308-1317.

Ai, D., G. Jiang, L. S. Kei and C. Li (2018). "Automatic Pixel-Level Pavement Crack Detection Using Information of Multi-Scale Neighborhoods." IEEE Access **6**: 24452-24463.

Azimi, M. and G. Pekcan (2019). "Structural Health Monitoring Using Extremely-compressed Data through Deep Learning." Computer-Aided Civil and Infrastructure Engineering.

Bang, S., S. Park, H. Kim and H. Kim (2019). "Encoder–decoder network for pixel-level road crack detection in black-box images." Computer‐Aided Civil Infrastructure Engineering **34**(8): 713-727.

Cha, Y. J., W. Choi and O. Büyüköztürk (2017). "Deep learning‐based crack damage detection using convolutional neural networks." Computer‐Aided Civil Infrastructure Engineering **32**(5): 361-378.

Cha, Y. J., W. Choi, G. Suh, S. Mahmoudkhani and O. Büyüköztürk (2018). "Autonomous structural visual inspection using region‐based deep learning for detecting multiple damage types." Computer‐Aided Civil Infrastructure Engineering **33**(9): 731-747.

Chan, R., M. Rottmann, F. Hüger, P. Schlicht and H. Gottschalk (2019). "Application of Decision Rules for Handling Class Imbalance in Semantic Segmentation." arXiv preprint arXiv:.08394.

Chen, X., R. Girshick, K. He and P. Dollár (2019). "Tensormask: A foundation for dense object segmentation." arXiv preprint arXiv:.12174.

Chollet, F. (2015). "Keras." from https://github.com/keras-team/keras.

Deng, J., W. Dong, R. Socher, L.-J. Li, K. Li and L. Fei-Fei (2009). Imagenet: A large-scale hierarchical image database. Computer Vision and Pattern Recognition, 2009. CVPR 2009. IEEE Conference on, Ieee.

Deng, J., Y. Lu and V. C.-S. Lee (2019). "Concrete crack detection with handwriting script interferences using faster region-based convolutional neural network." Computer-Aided Civil and Infrastructure Engineering **1**(17).

Ebrahimkhanlou, A., B. Dubuc and S. Salamone (2019). "A generalizable deep learning framework for localizing and characterizing acoustic emission sources in riveted metallic panels." Mechanical Systems and Signal Processing **130**: 248-272.

Eigen, D. and R. Fergus (2015). Predicting depth, surface normals and semantic labels with a common multi-scale convolutional architecture. Proceedings of the IEEE International Conference on Computer Vision.

Gal, Y. (2016). Uncertainty in deep learning, PhD thesis, University of Cambridge.

Gal, Y. and Z. Ghahramani (2015). "Bayesian convolutional neural networks with Bernoulli approximate variational inference." arXiv preprint arXiv:.02158.

Gal, Y. and Z. Ghahramani (2016). Dropout as a bayesian approximation: Representing model uncertainty in deep learning. international conference on machine learning.

Gao, Y. and K. Mosalam (2018). Structural ImageNet and PEER Hub ImageNet Challenge 2018.

Gao, Y. and K. M. Mosalam (2018). "Deep transfer learning for image‐based structural damage recognition." Computer‐Aided Civil and Infrastructure Engineering **33**(9): 748-768.

Goodfellow, I., Y. Bengio and A. Courville (2016). Deep learning, MIT press.

Graves, A. (2011). Practical variational inference for neural networks. Advances in neural information processing systems.

Hoskere, V., Y. Narazaki, T. Hoang and B. Spencer Jr (2018). "Vision-based Structural Inspection using Multiscale Deep Convolutional Neural Networks." arXiv preprint arXiv:.01055.

Huang, G., Z. Liu, L. Van Der Maaten and K. Q. Weinberger (2017). Densely connected convolutional networks. Proceedings of the IEEE conference on computer vision and pattern recognition.

Jégou, S., M. Drozdzal, D. Vazquez, A. Romero and Y. Bengio (2017). The one hundred layers tiramisu: Fully convolutional densenets for semantic segmentation. Proceedings of the IEEE Conference on Computer Vision and Pattern Recognition Workshops.

Karypidis, D. F., C. Berrocal, R. Rempling, G. Granath and P. J. I. Simonsson (2019). "Structural Health Monitoring of RC structures using optic fiber strain measurements: a deep learning approach."

Kendall, A., V. Badrinarayanan and R. Cipolla (2015). "Bayesian segnet: Model uncertainty in deep convolutional encoder-decoder architectures for scene understanding." arXiv preprint: arXiv:.00561.

Kendall, A. and Y. Gal (2017). What uncertainties do we need in bayesian deep learning for computer vision? Advances in neural information processing systems.

Khodabandehlou, H., G. Pekcan and M. S. Fadali (2019). "Vibration-based structural condition assessment using convolution neural networks." Structural Control and Health Monitoring **26**(2): e2308.

Koller, D. and N. Friedman (2009). Probabilistic graphical models: principles and techniques, MIT press.

Li, R., Y. Yuan, W. Zhang and Y. Yuan (2018). "Unified Vision‐Based Methodology for Simultaneous Concrete Defect Detection and Geolocalization." Computer‐Aided Civil and Infrastructure Engineering.

Li, S., X. Zhao and G. Zhou (2019). "Automatic pixel-level multiple damage detection of concrete structure using fully convolutional network." Computer‐Aided Civil Infrastructure Engineering **34**(7): 616-634.

Liang, X. (2019). "Image-based post-disaster inspection of reinforced concrete bridge systems using deep learning with Bayesian optimization." Computer‐Aided Civil Infrastructure Engineering **34**(5): 415-430.

Liang, X., Zheng, M., & Zhang, F. (2018). A scalable model-based learning algorithm with application to UAVs. IEEE control systems letters, 2(4), 839-844.

Liu, Y. F., X. Nie, J. S. Fan and X. G. Liu (2019). "Image‐based crack assessment of bridge piers using unmanned aerial vehicles and three‐dimensional scene reconstruction." Computer‐Aided Civil Infrastructure Engineering.

Long, J., E. Shelhamer and T. Darrell (2015). Fully convolutional networks for semantic segmentation. Proceedings of the IEEE conference on computer vision and pattern recognition.

Maeda, H., Y. Sekimoto, T. Seto, T. Kashiyama and H. J. C. A. C. Omata (2018). "Road damage detection and classification using deep neural networks with smartphone images." Computer‐Aided Civil Infrastructure Engineering **33**(12): 1127-1141.

Mathworks (2019). Image Labeler. Massachusetts, United States.

McAllister, R., Y. Gal, A. Kendall, M. Van Der Wilk, A. Shah, R. Cipolla and A. V. Weller (2017). Concrete problems for autonomous vehicle safety: Advantages of bayesian deep learning, International Joint Conferences on Artificial Intelligence, Inc.

Narazaki, Y., V. Hoskere, T. A. Hoang, Y. Fujino, A. Sakurai and B. F. Spencer Jr (2019). "Vision-based automated bridge component recognition with high-level scene consistency." Computer-Aided Civil and Infrastructure Engineering.

Oh, B. K., B. Glisic, Y. Kim and H. S. Park (2019). "Convolutional neural network-based wind-induced response estimation model for tall buildings." Computer-Aided Civil and Infrastructure Engineering **34**(10): 843-858.

Rafiei, M. H. and H. Adeli (2018). "A novel unsupervised deep learning model for global and local health condition assessment of structures." Engineering Structures **156**: 598-607.

Rafiei, M. H., W. H. Khushefati, R. Demirboga and H. Adeli (2017). "Supervised Deep Restricted Boltzmann Machine for Estimation of Concrete." ACI Materials Journal **114**(2).

Ronneberger, O., P. Fischer and T. Brox (2015). U-net: Convolutional networks for biomedical image segmentation. International Conference on Medical image computing and computer-assisted intervention, Springer.

Sajedi, S. O., & Liang, X. (2019, August). A convolutional cost-sensitive crack localization algorithm for automated and reliable RC bridge inspection. In Risk-Based Bridge Engineering: Proceedings of the 10th New York City Bridge Conference, August 26-27, 2019, New York City, USA (p. 229). CRC Press.

Sajedi, S. O., & Liang, X. (2020). A data-driven framework for near real-time and robust damage diagnosis of building structures. Structural Control and Health Monitoring, 27(3), e2488.

Sajedi, S. O., & Liang, X. (2019). Intensity-based feature selection for near real-time damage diagnosis of building structures. arXiv preprint arXiv:1910.11240.

Sajedi, S. O., & Liang, X. (2019). Vibration-based semantic damage segmentation for large-scale structural health monitoring. Computer-Aided Civil and Infrastructure Engineering.

Shi, Y., L. Cui, Z. Qi, F. Meng and Z. Chen (2016). "Automatic Road Crack Detection Using Random Structured Forests." IEEE Transactions on Intelligent Transportation Systems **17**(12): 3434-3445.

Spencer, B. F., V. Hoskere and Y. Narazaki (2019). "Advances in Computer Vision-Based Civil Infrastructure Inspection and Monitoring." Engineering **5**(2): 199-222.

Srivastava, N., G. Hinton, A. Krizhevsky, I. Sutskever and R. Salakhutdinov (2014). "Dropout: a simple way to prevent neural networks from overfitting." The journal of machine learning research **15**(1): 1929-1958.

Tong, Z., J. Gao, A. Sha, L. Hu and S. Li (2018). "Convolutional neural network for asphalt pavement surface texture analysis."

Computer‐Aided Civil Infrastructure Engineering **33**(12): 1056-1072.

Xue, Y. and Y. Li (2018). "A fast detection method via region‐based fully convolutional neural networks for shield tunnel lining defects." Computer‐Aided Civil Infrastructure Engineering **33**(8): 638-654.

Yang, X., H. Li, Y. Yu, X. Luo, T. Huang and X. Yang (2018). "Automatic Pixel-Level Crack Detection and Measurement Using Fully Convolutional Network." Computer‐Aided Civil and Infrastructure Engineering **33**(12): 1090-1109.

Zhang, A., K. C. P. Wang, B. Li, E. Yang, X. Dai, Y. Peng, Y. Fei, Y. Liu, J. Q. Li and C. Chen (2017). "Automated Pixel-Level Pavement Crack Detection on 3D Asphalt Surfaces Using a Deep-Learning Network." **32**(10): 805-819.

Zhang, C., C. c. Chang and M. Jamshidi (2019). "Concrete bridge surface damage detection using a single‐stage detector." Computer‐Aided Civil Infrastructure Engineering.

Zhang, X., D. Rajan and B. Story (2019). "Concrete crack detection using context‐aware deep semantic segmentation network." Computer‐Aided Civil Infrastructure Engineering.

Zheng, M., Chen, Z., & Liang, X. (2019, August). A Preliminary Study on a Physical Model Oriented Learning Algorithm With Application to UAVs. In ASME 2019 Dynamic Systems and Control Conference. American Society of Mechanical Engineers Digital Collection.